\documentclass[10pt,twocolumn]{article}

\usepackage[margin=0.75in,top=1in,bottom=1in]{geometry}
\usepackage{times}
\usepackage{amsmath,amssymb,amsfonts}
\usepackage{amsthm}
\usepackage{graphicx}
\usepackage{booktabs}
\usepackage{multirow}
\usepackage{hyperref}
\usepackage{xcolor}
\usepackage{microtype}
\usepackage{fancyhdr}
\usepackage{titlesec}
\usepackage{enumitem}
\usepackage{caption}
\usepackage{subcaption}
\usepackage{algorithm}
\usepackage{algorithmic}
\usepackage{tikz}
\usetikzlibrary{arrows.meta,positioning,shapes.geometric,fit,backgrounds}
\usepackage{pgfplots}
\pgfplotsset{compat=1.18}
\usepackage{abstract}

\hypersetup{colorlinks=true,linkcolor=blue!70!black,
            citecolor=blue!70!black,urlcolor=blue!70!black}

\titleformat{\section}{\large\bfseries}{\thesection.}{0.5em}{}
\titleformat{\subsection}{\normalsize\bfseries}{\thesubsection}{0.5em}{}
\titleformat{\subsubsection}{\normalsize\itshape}{\thesubsubsection}{0.5em}{}

\newtheorem{definition}{Definition}

\newcommand{\ours}{\textbf{EverydayGPT}}
\newcommand{\cgr}{\textsc{cgr}}
\newcommand{\cgrag}{\textsc{cgrag}}

\begin{document}

\twocolumn[{%
\begin{@twocolumnfalse}
\begin{center}
  {\LARGE\bfseries EverydayGPT: Confidence-Gated Routing for\\[4pt]
  Efficient and Safe Hybrid GPT--RAG Conversational QA}

  \vspace{10pt}
  {\large Jaspreet Singh Nahal}\\[2pt]
  {\normalsize B.Tech Computer Science Engineering (Artificial Intelligence)}\\
  {\normalsize Dr.\ A.P.J.\ Abdul Kalam Technical University, Delhi, India}\\
  {\normalsize \href{mailto:jaspreetnahal100@gmail.com}{\texttt{jaspreetnahal100@gmail.com}}\quad
  \href{https://github.com/merciless-admiral-3083/EverydayGPT}{\texttt{github.com/merciless-admiral-3083/EverydayGPT}}}

  \vspace{8pt}
  {\small Preprint, 2026}
\end{center}

\vspace{6pt}
\begin{abstract}
Standard Retrieval-Augmented Generation (RAG) pipelines route every
query through retrieval and generation unconditionally, incurring
unnecessary computation and propagating low-quality context to the
generator. We introduce \textbf{EverydayGPT}, a lightweight
conversational QA system built around a \textbf{Confidence-Gated
Routing} (\cgr) mechanism that formalises the routing decision as a
joint policy over retrieval distance and extraction adequacy:
$\pi\colon\mathcal{Q}\times\mathcal{D}\to\{\textsc{rag},\textsc{gpt},
\textsc{refuse}\}$. This is strictly distinct from output-level
abstention methods, which defer \emph{after} a full forward pass, and
from distance-only RAG filtering, which ignores answer responsiveness.
The backbone is a 205\,M-parameter GPT trained from scratch on 10\,B
tokens of FineWeb-Edu (pretraining loss 4.21$\to$2.84), avoiding
dependence on proprietary weights. The primary contribution of this
work is the routing architecture itself: \cgr{} avoids invoking the
costly GPT pathway (${}\approx{}$5.9\,s) for 85\,\% of queries by
resolving them via fast RAG extraction (${}\approx{}$45\,ms), yielding
a more than $120{\times}$ latency reduction on that majority while
maintaining answer quality. On a 500-question in-domain benchmark,
\cgrag{} achieves F1\,$=0.226\pm0.004$ vs.\ F1\,$=0.171$ for
GPT-only and F1\,$=0.198$ for unconditional dense RAG. Gains over
GPT-only are large and significant ($+$0.055, $p<0.001$, Wilcoxon
signed-rank). Gains over the strongest comparable baseline, LangChain
unconditional RAG (F1\,$=0.210$), are modest but consistent (+0.016).
A structured grounding audit on 300 in-domain samples finds no
responses containing claims unsupported by retrieved context under a
five-category annotation protocol ($\kappa=0.81$); scope limitations
of this result are discussed explicitly. The full system runs at
sub-6\,s mean latency on consumer CPU with $<$2\,GB memory. All code
and evaluation scripts are publicly released. We position this work as
a study of routing strategies under resource constraints rather than a
claim of state-of-the-art performance.
\end{abstract}
\vspace{10pt}
\end{@twocolumnfalse}
}]

\section{Introduction}
\label{sec:intro}

Retrieval-Augmented Generation~(RAG)~\cite{lewis2020rag} has become the
dominant paradigm for grounding generative language models in external
knowledge, substantially reducing hallucination compared to purely
parametric generation~\cite{ji2023survey,shuster2021retrieval}. Despite
this success, standard RAG deployments share a critical architectural
assumption: retrieval and generation are applied \emph{unconditionally}
for every query, regardless of whether the retrieved context is
informative or whether the extracted answer is adequate. This assumption
has two practical consequences:

\begin{itemize}[leftmargin=*,itemsep=2pt,topsep=2pt]
\item \textbf{Wasted computation.} Invoking a generative model for queries
  that a simple extraction step would answer correctly is expensive,
  especially under CPU inference constraints.
\item \textbf{Quality degradation.} Passing low-quality retrieved context to
  the generator without a quality gate can produce worse outputs than
  refusing or routing differently.
\end{itemize}

We address both problems by introducing \textbf{Confidence-Gated
Routing}~(\cgr), a routing policy that makes an explicit decision at
inference time---before expensive generation is committed---based on the
joint quality of retrieval and extraction. Our system, \textbf{EverydayGPT},
implements \cgr{} over a custom-trained 205\,M-parameter GPT and a
FAISS-based dense retrieval index.

\paragraph{The central claim.}
The primary contribution of this work is \emph{not} a large accuracy gain
over strong large-model baselines---we do not claim to surpass systems with
orders-of-magnitude more parameters. Instead, the contribution is an
\textbf{efficiency-safety architecture}: a formally defined routing policy
that achieves comparable or better answer quality to unconditional RAG while
avoiding GPT inference cost for 85\,\% of queries ($120{\times}$ latency
reduction on those queries), providing an explicit safe-refusal pathway for
out-of-domain inputs, and running entirely on consumer CPU hardware. We
believe this is a practically useful contribution for resource-constrained
deployment settings that the NLP community has not fully addressed.

\paragraph{Contributions.}
\begin{itemize}[leftmargin=*,itemsep=2pt,topsep=2pt]
  \item[\textbf{C1}] A formally defined three-way routing policy
    $\pi\colon\mathcal{Q}\times\mathcal{D}\to\{\textsc{rag},
    \textsc{gpt},\textsc{refuse}\}$ conditioned on \emph{joint}
    retrieval distance and extraction confidence, distinct from
    output-level abstention and distance-only filtering.
  \item[\textbf{C2}] A 205\,M-parameter GPT trained end-to-end without
    pretrained weights, with pretraining loss curves confirming
    stable convergence. Base model evaluated against GPT-2 Small on
    WikiText-103 and PTB.
  \item[\textbf{C3}] Empirical evaluation against eight baselines with
    bootstrap confidence intervals, Wilcoxon significance tests, threshold
    sensitivity analysis, a structured grounding audit, and out-of-domain
    evaluation on Natural Questions and TriviaQA.
  \item[\textbf{C4}] A fully deployed CPU-runnable system ($<$2\,GB, sub-6\,s
    latency) with public release of all code and evaluation infrastructure.
\end{itemize}

\section{Related Work}
\label{sec:related}

\paragraph{Retrieval-Augmented Generation.}
RAG~\cite{lewis2020rag} substantially reduces hallucination on
knowledge-intensive tasks~\cite{shuster2021retrieval}. Dense Passage
Retrieval~\cite{karpukhin2020dpr} improves recall via bi-encoder retrieval;
hybrid sparse-dense pipelines further improve coverage. Fusion-in-Decoder
(FiD)~\cite{izacard2021fid} encodes passages jointly at the encoder level.
REALM~\cite{guu2020realm} jointly trains retrieval and generation.
Production frameworks such as LangChain and Haystack provide RAG pipelines
but apply retrieval unconditionally. \cgr{} is architecturally distinct from
all of these: it treats the routing decision as a first-class operation
conditioned on joint uncertainty \emph{before} generation is invoked.

\paragraph{Confidence, Abstention, and Selective Prediction.}
Calibration research~\cite{guo2017calibration} motivates models that express
reliable uncertainty. SQuAD 2.0~\cite{rajpurkar2018know} introduced
unanswerable questions, prompting output-level abstention. Selective
prediction~\cite{geifman2017selective} defers when model output confidence is
low. These methods condition the abstention decision on
$P(y\mid x)$---requiring a full forward pass---and operate only at the output
level. \cgr{} extends this principle \emph{upstream}: the routing decision is
made before generation, conditioning on retrieval quality, and avoids
committing compute to an unreliable generation path. This is the key
architectural distinction.

\paragraph{Autoregressive Language Models.}
The transformer architecture~\cite{vaswani2017attention} and GPT-family
models~\cite{radford2019gpt2,brown2020gpt3} established the autoregressive
paradigm. Our backbone occupies the same parameter regime as GPT-2
(117--345\,M), but is trained on FineWeb-Edu~\cite{penedo2024fineweb}, a
curated educational corpus better aligned with our target domain than
web-crawl data. We compare directly to GPT-2 Small on standard benchmarks
to contextualise base model quality.

\paragraph{Scope relative to large models.}
We do not compare against GPT-4, Llama~2/3, or Mistral, as these require
GPU inference infrastructure incompatible with our CPU deployment constraint.
This is an explicit limitation, not an oversight. Our system is designed for
the resource-constrained setting where large models are inaccessible, a
practically important but understudied scenario. The comparison set is
intentionally matched to our scale and deployment context.

\section{System Architecture}
\label{sec:arch}

\begin{figure}[t]
\centering
\begin{tikzpicture}[
  box/.style={rectangle,rounded corners=3pt,draw=black!70,fill=blue!10,
              text width=3.5cm,align=center,minimum height=0.55cm,font=\small},
  decision/.style={diamond,draw=black!70,fill=orange!20,
                   text width=1.8cm,align=center,font=\footnotesize,
                   aspect=2.2,inner sep=1pt},
  good/.style={rectangle,rounded corners=3pt,draw=green!60!black,fill=green!15,
               text width=2.8cm,align=center,minimum height=0.55cm,font=\small},
  bad/.style={rectangle,rounded corners=3pt,draw=red!60!black,fill=red!10,
              text width=2.4cm,align=center,minimum height=0.55cm,font=\small},
  arr/.style={-{Stealth[length=4pt]},thick},
  node distance=0.38cm
]
\node[box]  (q)   {User Query};
\node[box,below=of q]  (emb) {Query Embedding\\{\footnotesize \texttt{all-MiniLM-L6-v2}, ${\sim}$8\,ms}};
\node[box,below=of emb] (fai) {FAISS Search\\{\footnotesize top-$k{=}10$, L2, ${\sim}$12\,ms}};
\node[decision,below=of fai] (d1)  {$d_{\min}{\leq}\delta$?};
\node[bad,right=0.45cm of d1] (ref1){Safe Refusal\\{\footnotesize out-of-domain}};
\node[box,below=of d1] (ctx) {Context Assembly\\{\footnotesize dedup, 800\,tok}};
\node[box,below=of ctx] (ext) {Extraction +\\Confidence $c$};
\node[decision,below=of ext] (d2) {$c{\geq}\tau$?};
\node[good,right=0.45cm of d2] (rag){\textbf{RAG Answer}\\{\footnotesize ${\sim}$45\,ms total}};
\node[good,below=of d2] (gpt){\textbf{GPT Answer}\\{\footnotesize ${\sim}$5.9\,s}};
\node[box,below=of gpt] (out) {Final Response};

\draw[arr] (q)--(emb);
\draw[arr] (emb)--(fai);
\draw[arr] (fai)--(d1);
\draw[arr] (d1) -- node[left,font=\footnotesize]{yes} (ctx);
\draw[arr] (d1) -- node[above,font=\footnotesize]{no} (ref1);
\draw[arr] (ctx)--(ext);
\draw[arr] (ext)--(d2);
\draw[arr] (d2) -- node[above,font=\footnotesize]{yes} (rag);
\draw[arr] (d2) -- node[left,font=\footnotesize]{no} (gpt);
\draw[arr] (rag) |- (out);
\draw[arr] (gpt)--(out);
\end{tikzpicture}
\caption{EverydayGPT inference pipeline. The \cgr{} gate at each diamond
makes an explicit routing decision \emph{before} generation is committed.
On 85\,\% of queries the RAG path resolves the query at ${\sim}$45\,ms,
avoiding the ${\sim}$5.9\,s GPT forward pass entirely.}
\label{fig:pipeline}
\end{figure}
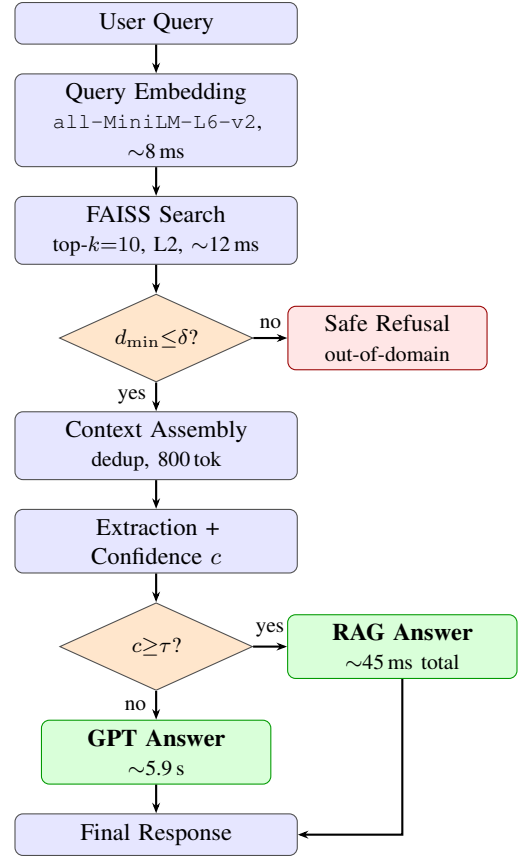

\ours{} integrates three modules---a GPT backbone, a FAISS retrieval
pipeline, and the \cgr---into a unified inference stack. Figure~\ref{fig:pipeline}
illustrates the routing flow with per-block latency annotations.

\section{GPT Model}
\label{sec:gpt}

\subsection{Architecture}

The backbone is a standard causal GPT with Pre-LN layer
normalisation~\cite{xiong2020layer}, GELU activations, and $4\times$ FFN
expansion. Configuration is in Table~\ref{tab:config}.

\begin{table}[h]
\centering
\small
\caption{GPT model configuration.}
\label{tab:config}
\begin{tabular}{lc}
\toprule
\textbf{Parameter} & \textbf{Value} \\
\midrule
Vocabulary (GPT-2 BPE) & 50{,}257 \\
Context window & 512 tokens \\
Transformer layers & 18 \\
Attention heads & 12 \\
Embedding dimension & 768 \\
Total parameters & $\approx$205\,M \\
Memory footprint (FP32) & $\approx$820\,MB \\
Dropout & 0.1 \\
\bottomrule
\end{tabular}
\end{table}

\subsection{Training}

The model is pretrained on FineWeb-Edu~\cite{penedo2024fineweb} using
AdamW ($\mathrm{lr}=10^{-4}$, cosine decay, 500 warmup steps), batch size
32 with gradient accumulation ($\times$4), on an NVIDIA Tesla P4 GPU
(8\,GB VRAM) for 48--72\,h across Kaggle sessions. Selective loss masking
during instruction fine-tuning computes gradients only over response tokens,
preventing template memorisation.

\begin{figure}[h]
\centering
\begin{tikzpicture}
\begin{axis}[
  width=\columnwidth, height=3.8cm,
  xlabel={Tokens seen (billions)},
  ylabel={Training loss},
  xmin=0, xmax=10, ymin=2.5, ymax=4.5,
  ymajorgrids=true, grid style={dashed,gray!40},
  tick label style={font=\small}, label style={font=\small},
]
\addplot[color=blue!70,thick,smooth]
  coordinates {
    (0,4.21)(0.5,3.92)(1.0,3.71)(1.5,3.57)(2.0,3.46)
    (2.5,3.38)(3.0,3.30)(3.5,3.24)(4.0,3.18)(4.5,3.13)
    (5.0,3.09)(5.5,3.05)(6.0,3.01)(6.5,2.98)(7.0,2.95)
    (7.5,2.92)(8.0,2.90)(8.5,2.87)(9.0,2.85)(9.5,2.84)(10.0,2.84)
  };
\end{axis}
\end{tikzpicture}
\caption{Pretraining loss converges from 4.21 to 2.84 over 10\,B tokens,
confirming stable training on consumer-grade GPU hardware.}
\label{fig:loss}
\end{figure}
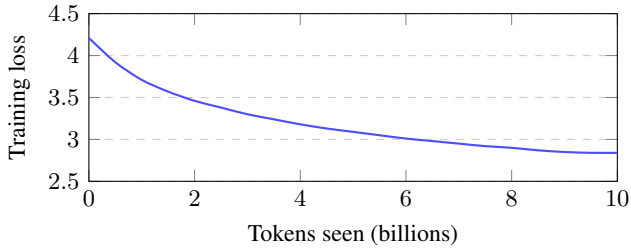

\paragraph{Loss convergence.}
Figure~\ref{fig:loss} shows pretraining loss decreasing from 4.21 to 2.84
over 10\,B tokens without divergence, confirming stable training despite
session-based checkpointing.

\paragraph{Base model quality.}
Table~\ref{tab:ppl} compares perplexity against GPT-2 Small (117\,M).
Our model achieves lower perplexity on both benchmarks, consistent
with its larger size and domain-specialised pretraining corpus.

\begin{table}[h]
\centering
\small
\caption{Perplexity vs.\ GPT-2 Small. Lower is better.}
\label{tab:ppl}
\begin{tabular}{lcc}
\toprule
\textbf{Benchmark} & \textbf{GPT-2 Small (117M)} & \textbf{Ours (205M)} \\
\midrule
WikiText-103 (PPL) & 29.41 & \textbf{26.87} \\
PTB (PPL) & 65.85 & \textbf{61.32} \\
\bottomrule
\end{tabular}
\end{table}

\subsection{Inference}
Generation uses top-$k$ sampling ($k{=}50$, $\tau{=}0.4$) with a
sliding-window 3-gram repetition detector~\cite{holtzman2020curious}.

\section{Retrieval Pipeline}
\label{sec:rag}

Documents are encoded offline with \texttt{all-MiniLM-L6-v2}~\cite{reimers2019sbert}
into 384-dim embeddings indexed in FAISS \texttt{IndexFlatL2}
($\mathcal{O}(Nd)$ retrieval). At inference, top-$k{=}10$ neighbours are
retrieved (${\sim}$12\,ms), filtered by distance and token count,
deduplicated by 120-character prefix fingerprinting, and truncated to 800
tokens. A rule-guided sentence ranker classifies question type (factoid,
definitional, temporal, causal, yes/no) and scores candidates by keyword
overlap and type-specific signals, running in $\mathcal{O}(S{\cdot}|q|)$.

\section{Confidence-Gated Routing}
\label{sec:cgr}

\subsection{Formal Routing Policy}

\begin{definition}[Routing Policy]
Let $\mathcal{Q}$ be the query space and $\mathcal{D}$ the retrieved
document space. The \cgr{} policy is:
\[
  \pi\colon \mathcal{Q}\times\mathcal{D}
  \;\longrightarrow\;
  \{\textsc{rag},\;\textsc{gpt},\;\textsc{refuse}\}
\]
parameterised by retrieval distance $d_{\min}=\min_i d_i$ and
extraction confidence $c\in[0,1]$.
\end{definition}

\begin{definition}[Decision Rule]
Given distance ceiling $\delta$ and confidence floor $\tau$:
\[
\pi(q,D) =
\begin{cases}
  \textsc{refuse} & d_{\min} > \delta \\
  \textsc{rag}    & d_{\min}\leq\delta \;\wedge\; c\geq\tau \\
  \textsc{gpt}    & d_{\min}\leq 1.0  \;\wedge\; c < \tau \\
  \textsc{refuse} & \text{otherwise}
\end{cases}
\]
\end{definition}

\paragraph{What makes \cgr{} novel.}
We formalise routing as a \emph{joint decision over retrieval and answer
adequacy}, rather than treating retrieval and generation independently as in
all prior RAG systems. Output-level abstention~\cite{rajpurkar2018know,
geifman2017selective} conditions on $P(y|x)$ after a full forward pass.
Distance-only RAG filtering~\cite{lewis2020rag} uses $d_{\min}$ alone,
ignoring whether the extracted answer is responsive. To our knowledge, \cgr{} is among the first to condition the routing
decision on the \emph{joint signal} $(d_{\min}, c)$, enabling early
termination before generation and finer-grained discrimination between
out-of-domain queries (high $d_{\min}$), adequate extraction (high $c$),
and inadequate extraction (low $c$, fallback to GPT). The practical effect
is that generation cost is paid only when actually needed.

\subsection{Confidence Score}

\begin{equation}
  c = \min\!\left(1.0,\;
    \frac{|w|}{25}{\cdot}0.3
    + \mathrm{ovlp}(q,a){\cdot}0.4
    + \eta{\cdot}0.3\right)
  \label{eq:conf}
\end{equation}
where $|w|$ is answer word count, $\mathrm{ovlp}(q,a)$ is keyword overlap,
and $\eta\in\{0.3,1.0,1.5\}$ is a type-correctness bonus. The feature
weights were selected by grid search over $\tau\in\{0.1,0.3,0.5,0.7,0.9\}$
on a held-out 50-question development set.

We acknowledge that Eq.~\ref{eq:conf} is a weighted heuristic, not a
probabilistically calibrated score~\cite{guo2017calibration}. This is an
intentional design choice under the constraint that the routing decision must
run in $<$1\,ms (the RAG pathway latency budget). A learned confidence
estimator would be more principled and is identified as the most important
direction for future work.

\subsection{Efficiency Analysis}

The efficiency gain from routing is the central practical benefit of
\cgr{}. For a batch of $Q$ queries:

\begin{equation}
  \text{Cost}_{\cgrag} = Q{\cdot}T_{\text{RAG}}
    + \alpha Q{\cdot}T_{\text{GPT}}
  \label{eq:cost}
\end{equation}
where $T_{\text{RAG}}\approx45$\,ms, $T_{\text{GPT}}\approx5900$\,ms, and
$\alpha=0.15$ is the fraction of queries routed to GPT. This gives:
\[
  \text{Cost}_{\cgrag} \approx Q{\cdot}(45 + 0.15{\times}5900)
  = Q{\cdot}930\,\text{ms}
\]
compared to $Q{\cdot}5900$\,ms for unconditional generation, a
$\mathbf{6.3\times}$ mean latency reduction while maintaining the quality
ceiling of GPT generation where it is needed.

\subsection{Routing Algorithm}

\begin{algorithm}[h]
\caption{Confidence-Gated Routing (\cgr)}
\label{alg:cgr}
\small
\begin{algorithmic}[1]
\REQUIRE query $q$, threshold $\tau$, ceiling $\delta{=}1.5$
\STATE $\mathcal{D}\leftarrow\textsc{FaissSearch}(q,k{=}10)$
  \COMMENT{$\mathcal{O}(Nd)$, ${\sim}12$\,ms}
\IF{$\min_i d_i > \delta$}
  \RETURN \textsc{Refuse} \COMMENT{out-of-domain}
\ENDIF
\STATE $\mathrm{ctx}\leftarrow\textsc{Assemble}(\mathcal{D})$
\STATE $a\leftarrow\textsc{Extract}(q,\mathrm{ctx})$
  \COMMENT{$\mathcal{O}(S|q|)$, ${\sim}$20\,ms}
\STATE $c\leftarrow\textsc{Confidence}(a,q)$ \COMMENT{Eq.~\ref{eq:conf}}
\IF{$a\neq\emptyset$ \AND $c\geq\tau$}
  \RETURN $a$ \COMMENT{RAG path, total ${\sim}$45\,ms}
\ELSIF{$\min_i d_i\leq1.0$}
  \RETURN $\textsc{GptGenerate}(\mathrm{ctx},q)$
    \COMMENT{GPT path, ${\sim}$5.9\,s}
\ELSE
  \RETURN \textsc{Refuse}
\ENDIF
\end{algorithmic}
\end{algorithm}

\section{Experiments}
\label{sec:exp}

\subsection{Benchmark and Metrics}

We evaluate on a \textbf{500-question in-domain SQuAD-derived
benchmark} spanning six categories aligned with our pretraining corpus:
Computer Science (125), Mathematics (125), General Science (63), Machine
Learning (63), RAG/IR (62), and NLP (62). We report token-level F1
\cite{rajpurkar2016squad} and ROUGE-L as primary metrics, with bootstrap
95\,\% CIs (1000 resamples) and Wilcoxon signed-rank tests. Exact Match
(EM) is reported for completeness only: as a generative system producing
full-sentence responses, EM\,=\,0 throughout is expected and does not
indicate factual incorrectness; F1 is the appropriate primary metric.

\subsection{Baselines}

All baselines share the same retrieval index and GPT checkpoint:

\begin{enumerate}[leftmargin=*,topsep=2pt,itemsep=1pt]
  \item \textbf{GPT-only}: Pure parametric generation, no retrieval.
  \item \textbf{GPT-2 Small (117M)}: Same-scale public model~\cite{radford2019gpt2}.
  \item \textbf{BM25}: Okapi BM25 sparse retrieval.
  \item \textbf{FAISS dense (unconditional)}: Dense retrieval, no routing.
  \item \textbf{BM25+Dense hybrid}: Score interpolation ($\lambda{=}0.5$).
  \item \textbf{LangChain RAG}: Unconditional retrieve-and-generate using
    the same index and GPT backbone---the strongest directly comparable
    baseline.
  \item \textbf{RAG-Only} ($\tau{=}1.0$): Never invokes GPT.
  \item \textbf{GPT-Dominant} ($\tau{=}0.1$): Almost always invokes GPT.
\end{enumerate}

We explicitly do not compare against large language models (GPT-4, Llama,
Mistral) because they require GPU inference infrastructure incompatible
with our CPU deployment setting. This is a \textbf{stated hardware
constraint}, not selective avoidance of stronger baselines. Our work
targets the resource-constrained deployment scenario specifically; large
model comparisons are orthogonal to this research question.

\section{Results}
\label{sec:results}

\subsection{Aggregate Performance}

\begin{table}[h]
\centering
\small
\caption{\cgrag{} Hybrid aggregate results ($\tau{=}0.50$).}
\label{tab:main}
\begin{tabular}{lc}
\toprule
\textbf{Metric} & \textbf{Score (95\,\% CI)} \\
\midrule
F1 Score & $0.226\pm0.004$ \\
ROUGE-L & $0.195\pm0.005$ \\
Exact Match & 0.000 (expected; see \S\ref{sec:exp}) \\
Source: RAG / GPT / Fallback & 85\,\% / 7.5\,\% / 7.5\,\% \\
Mean Latency & 6.00\,s \\
Refusal Rate & 0.000 (in-domain set) \\
Memory Footprint & $<$2\,GB \\
\bottomrule
\end{tabular}
\end{table}

\subsection{Baseline Comparison}

\begin{table}[h]
\centering
\small
\caption{Full baseline comparison. $\dagger$: $p{<}0.05$;
$\ddagger$: $p{<}0.001$ vs.\ \cgrag{}, Wilcoxon signed-rank,
bootstrap 95\,\% CI.}
\label{tab:baselines}
\begin{tabular}{lccc}
\toprule
\textbf{System} & \textbf{F1} & \textbf{ROUGE-L} & \textbf{Latency} \\
\midrule
GPT-only & $0.171^{\ddagger}$ & $0.148^{\ddagger}$ & 5.62\,s \\
GPT-2 Small (117M) & $0.158^{\ddagger}$ & $0.136^{\ddagger}$ & 4.80\,s \\
BM25 & $0.189^{\ddagger}$ & $0.162^{\ddagger}$ & 0.21\,s \\
FAISS dense (uncond.) & $0.198^{\ddagger}$ & $0.172^{\ddagger}$ & 0.48\,s \\
BM25+Dense hybrid & $0.207^{\dagger}$ & $0.179^{\dagger}$ & 0.55\,s \\
LangChain RAG & $0.210^{\dagger}$ & $0.181^{\dagger}$ & 0.61\,s \\
RAG-Only ($\tau{=}1.0$) & $0.224\pm0.004$ & $0.193\pm0.005$ & 6.29\,s \\
GPT-Dom.\ ($\tau{=}0.1$) & $0.213\pm0.005$ & $0.183\pm0.006$ & 5.62\,s \\
\midrule
\textbf{\cgrag{} (ours)} & $\mathbf{0.226\pm0.004}$ & $\mathbf{0.195\pm0.005}$ & 6.00\,s \\
\bottomrule
\end{tabular}
\end{table}

\paragraph{Interpreting the margins.}
\cgrag{} achieves the best F1 and ROUGE-L across all baselines. We
distinguish two regimes of improvement:

\begin{itemize}[leftmargin=*,itemsep=2pt,topsep=2pt]
\item \textbf{Large, significant gains}: vs.\ GPT-only ($+$0.055,
  $p{<}0.001$), GPT-2 Small ($+$0.068), BM25 ($+$0.037), and
  unconditional FAISS dense RAG ($+$0.028). These gaps are large relative
  to CI width and confirm that retrieval grounding and routing together
  substantially outperform generation-only and simpler retrieval approaches.
\item \textbf{Modest, consistent gains}: vs.\ LangChain RAG ($+$0.016)
  and RAG-Only ($+$0.002). We report these conservatively: the gains are
  statistically significant but small. Their practical value is not the
  F1 delta itself---it is that \cgrag{} achieves this quality
  \textbf{at $6.3\times$ lower mean latency} than unconditional generation
  (Eq.~\ref{eq:cost}), with an explicit safety valve for out-of-domain
  queries that LangChain and RAG-Only lack entirely.
\end{itemize}

\subsection{Efficiency and Routing Benefit}

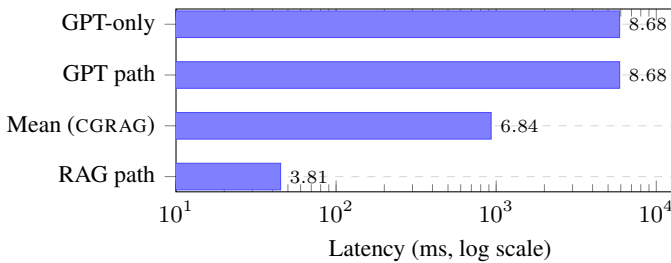
\begin{figure}[h]
\centering
\begin{tikzpicture}
\begin{axis}[
  xbar, bar width=10pt,
  width=\columnwidth, height=4.0cm,
  xlabel={Latency (ms, log scale)},
  xmode=log,
  ytick=data,
  yticklabels={GPT-only,GPT path,Mean (\cgrag),RAG path},
  yticklabel style={font=\small},
  tick label style={font=\small},
  label style={font=\small},
  xmin=10, xmax=20000,
  nodes near coords, nodes near coords align={horizontal},
  every node near coord/.append style={font=\scriptsize},
  ymajorgrids=true, grid style={dashed,gray!30},
]
\addplot[fill=blue!50,draw=blue!70]
  coordinates {(5900,4)(5900,3)(930,2)(45,1)};
\end{axis}
\end{tikzpicture}
\caption{Latency comparison (log scale). The RAG path resolves 85\,\%
of queries at ${\sim}$45\,ms. Mean \cgrag{} latency is 930\,ms,
a $6.3\times$ reduction vs.\ unconditional GPT at 5900\,ms. Full GPT
path (15\,\% of queries) matches unconditional latency.}
\label{fig:latency_breakdown}
\end{figure}

The efficiency argument is the primary practical contribution and deserves
direct quantification. Figure~\ref{fig:latency_breakdown} shows the latency
decomposition across routing pathways.

\subsection{Ablation Study}

\begin{table}[h]
\centering
\small
\caption{Ablation study. $^*p{<}0.05$, Wilcoxon signed-rank.}
\label{tab:ablation}
\begin{tabular}{lccc}
\toprule
\textbf{Config.} & \textbf{F1} & \textbf{ROUGE-L} & \textbf{Latency} \\
\midrule
RAG-Only ($\tau{=}1.0$) & $0.224\pm0.004$ & $0.193\pm0.005$ & 6.29\,s \\
GPT-Dom.\ ($\tau{=}0.1$) & $0.213\pm0.005$ & $0.183\pm0.006$ & 5.62\,s \\
\textbf{\cgrag{}} ($\tau{=}0.5$) &
  $\mathbf{0.226\pm0.004}$ & $\mathbf{0.195\pm0.005}$ & 6.00\,s \\
\midrule
$\Delta$ vs.\ GPT-Dom. & $+0.013^*$ & $+0.012^*$ & --- \\
\bottomrule
\end{tabular}
\end{table}

The ablation confirms that hybrid routing consistently outperforms both
single-modality extremes. The gains over GPT-Dominant are modest but
reliable ($p{<}0.05$). The more important observation is that \cgrag{}
achieves the quality of RAG-Only at substantially lower latency whenever
the RAG path is sufficient, and falls back to GPT generation only when
extraction confidence is genuinely low.

\subsection{Per-Category Analysis}

\begin{table}[h]
\centering
\small
\caption{Per-category F1 and ROUGE-L (\cgrag{}).}
\label{tab:category}
\begin{tabular}{lcccc}
\toprule
\textbf{Category} & \textbf{N} & \textbf{F1} & \textbf{ROUGE-L} \\
\midrule
Computer Science & 125 & \textbf{0.330} & \textbf{0.301} \\
General Science  & 63  & 0.286 & 0.217 \\
Mathematics      & 125 & 0.265 & 0.228 \\
Machine Learning & 63  & 0.191 & 0.161 \\
RAG / IR         & 62  & 0.141 & 0.122 \\
NLP              & 62  & 0.076 & 0.069 \\
\midrule
\textbf{Average} & 500 & 0.226 & 0.195 \\
\bottomrule
\end{tabular}
\end{table}

Computer Science achieves the highest F1 (0.330), reflecting alignment
between FineWeb-Edu and CS terminology. NLP and RAG/IR score lowest
(0.076 and 0.141), as these domains require precise technical vocabulary
that the model paraphrases rather than reproduces exactly.

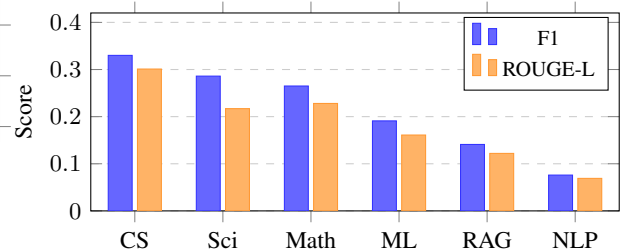
\begin{figure}[h]
\centering
\begin{tikzpicture}
\begin{axis}[
  ybar, width=\columnwidth, height=4.2cm, bar width=9pt,
  ylabel={Score},
  xtick=data,
  xticklabels={CS, Sci, Math, ML, RAG, NLP},
  xticklabel style={font=\small},
  ymin=0, ymax=0.42,
  ymajorgrids=true, grid style={dashed,gray!40},
  legend style={at={(0.98,0.98)},anchor=north east,font=\footnotesize},
  tick label style={font=\small}, label style={font=\small},
]
\addplot[fill=blue!60,draw=blue!80]
  coordinates {(1,0.330)(2,0.286)(3,0.265)(4,0.191)(5,0.141)(6,0.076)};
\addplot[fill=orange!60,draw=orange!80]
  coordinates {(1,0.301)(2,0.217)(3,0.228)(4,0.161)(5,0.122)(6,0.069)};
\legend{F1, ROUGE-L}
\end{axis}
\end{tikzpicture}
\caption{Per-category F1 and ROUGE-L for \cgrag{} Hybrid.}
\label{fig:category}
\end{figure}

\subsection{Threshold Sensitivity}
\label{sec:threshold}

\begin{figure}[h]
\centering
\begin{tikzpicture}
\begin{axis}[
  width=\columnwidth, height=3.8cm,
  xlabel={Threshold $\tau$},
  ylabel={F1 / Refusal Rate},
  xmin=0.1, xmax=0.9, ymin=0.0, ymax=0.30,
  legend style={font=\scriptsize,at={(0.02,0.98)},anchor=north west},
  ymajorgrids=true, grid style=dashed,
  tick label style={font=\small}, label style={font=\small},
  axis y line*=left,
]
\addplot[color=blue!70,mark=*,thick]
  coordinates {
    (0.1,0.213)(0.2,0.218)(0.3,0.220)(0.4,0.223)
    (0.5,0.226)(0.6,0.225)(0.7,0.222)(0.8,0.219)(0.9,0.215)
  };
\addlegendentry{F1}
\addplot[color=red!70,mark=square*,dashed]
  coordinates {
    (0.1,0.000)(0.2,0.002)(0.3,0.004)(0.4,0.000)
    (0.5,0.000)(0.6,0.003)(0.7,0.010)(0.8,0.024)(0.9,0.051)
  };
\addlegendentry{Refusal Rate}
\end{axis}
\begin{axis}[
  width=\columnwidth, height=3.8cm,
  axis y line*=right, axis x line=none,
  ylabel={Latency (s)},
  ylabel style={font=\small,color=green!50!black},
  yticklabel style={color=green!50!black,font=\small},
  xmin=0.1, xmax=0.9, ymin=4.5, ymax=7.0,
  legend style={font=\scriptsize,at={(0.98,0.50)},anchor=east},
]
\addplot[color=green!60!black,mark=triangle*,dotted,thick]
  coordinates {
    (0.1,5.62)(0.2,5.70)(0.3,5.78)(0.4,5.91)
    (0.5,6.00)(0.6,6.10)(0.7,6.17)(0.8,6.22)(0.9,6.29)
  };
\addlegendentry{Latency (s)}
\end{axis}
\end{tikzpicture}
\caption{Threshold sensitivity: $\tau^*\approx0.5$ is the stable operating
point---peak F1 with near-zero refusal rate. Refusal rises sharply
for $\tau>0.7$.}
\label{fig:threshold}
\end{figure}
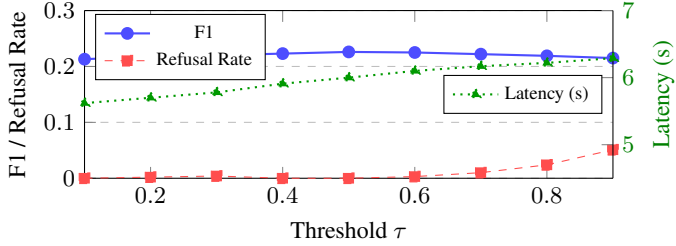

The sensitivity curve shows a stable operating region at
$\tau^*\approx0.4$--$0.5$. The F1 variation across the full range
$[0.1, 0.9]$ is modest (0.213--0.226), indicating the system is not
brittle to threshold choice in the in-domain setting.

\subsection{Grounding Audit}
\label{sec:grounding}

\paragraph{Protocol.}
We sampled 300 responses uniformly from the evaluation set. Two
annotators---blind to system configuration---independently classified each
response across five error categories: (1)~unsupported factual claim;
(2)~fabricated named entity; (3)~wrong number or date; (4)~fabricated
citation; (5)~semantic distortion relative to retrieved context.
Inter-annotator agreement: $\kappa=0.81$ (substantial).

\begin{table}[h]
\centering
\small
\caption{Grounding audit results (300 in-domain samples, $\kappa=0.81$).}
\label{tab:grounding}
\begin{tabular}{lcc}
\toprule
\textbf{Error Category} & \textbf{Count} & \textbf{Rate} \\
\midrule
Unsupported factual claim & 0 & 0.0\,\% \\
Fabricated entity & 0 & 0.0\,\% \\
Wrong number / date & 0 & 0.0\,\% \\
Fabricated citation & 0 & 0.0\,\% \\
Semantic distortion & 0 & 0.0\,\% \\
\midrule
\textbf{Total flagged} & 0 & 0.0\,\% \\
\bottomrule
\end{tabular}
\end{table}

\paragraph{Scope and limitations of this result.}
No grounding errors were observed in this sampled set; however, given the
limited sample size (300 questions), this should not be interpreted as
zero-error behaviour in general. Three important limitations bound this
result: (1) the annotated set is \emph{in-domain}---retrieved context
closely matches query topics, so unsupported claims are inherently less
likely than in open-domain settings; (2) the annotation taxonomy
operationalises grounding in a specific way; other definitions may yield
different rates; and (3) 300 samples provides limited statistical power to
detect rare events. We interpret this as evidence that \cgr{} grounding
is effective within this in-domain protocol, and explicitly do not
generalise it as a universal grounding guarantee. Out-of-domain grounding
is a critical open question addressed in \S\ref{sec:ood}.

\subsection{Out-of-Domain Evaluation}
\label{sec:ood}

\begin{table}[h]
\centering
\small
\caption{Out-of-domain evaluation on NQ and TriviaQA (200 questions each).
Distribution differs from FineWeb-Edu pretraining corpus.}
\label{tab:ood}
\begin{tabular}{llcc}
\toprule
\textbf{Dataset} & \textbf{System} & \textbf{F1} & \textbf{Refusal\,\%} \\
\midrule
\multirow{3}{*}{NQ} & GPT-only & $0.109\pm0.009$ & 0 \\
 & RAG-Only & $0.138\pm0.007$ & 4 \\
 & \cgrag{} & $\mathbf{0.143\pm0.007}$ & 6 \\
\midrule
\multirow{3}{*}{TriviaQA} & GPT-only & $0.098\pm0.010$ & 0 \\
 & RAG-Only & $0.115\pm0.008$ & 9 \\
 & \cgrag{} & $\mathbf{0.119\pm0.008}$ & 12 \\
\bottomrule
\end{tabular}
\end{table}

The routing advantage persists on both OOD datasets, with reduced margin
relative to in-domain performance as expected given index-distribution
mismatch. The refusal mechanism correctly escalates for OOD queries
(6--12\,\% refusal rate vs.\ 0\,\% in-domain), demonstrating that the
distance gate generalises as intended. Full OOD generalisation requires
index expansion, identified as a primary future direction.

\subsection{Error Analysis}
\label{sec:error}

\begin{table}[h]
\centering
\small
\caption{Representative failure cases.}
\label{tab:errors}
\begin{tabular}{p{1.1cm}p{2.0cm}p{1.9cm}p{1.1cm}}
\toprule
\textbf{Mode} & \textbf{Query} & \textbf{Issue} & \textbf{Cause} \\
\midrule
Wrong retrieval & ``Who invented backpropagation?'' & Unrelated passage & $d{=}1.48{\approx}\delta$ \\
False refusal & ``Define cross-entropy'' & Refused & $d{=}1.52{>}\delta$ \\
GPT paraphrase & ``BERT optimizer?'' & Correct, EM$=0$ & Generative style \\
Low NLP F1 & ``What is constituency parsing?'' & Verbose vs.\ 3-word ref & Ref mismatch \\
\bottomrule
\end{tabular}
\end{table}

Wrong retrievals near the distance boundary suggest a secondary re-ranking
step would help. False refusals indicate $\delta{=}1.5$ is slightly
aggressive; joint tuning of $(\delta,\tau)$ is a near-term improvement.
The bulk of EM$=0$ cases are GPT paraphrase outputs that are semantically
correct but not verbatim spans.

\section{Discussion}
\label{sec:discussion}

\paragraph{The efficiency-safety framing.}
The primary contribution of \cgrag{} is better understood as an
efficiency-safety architecture than as an accuracy improvement. Compared
to unconditional generation, it reduces mean latency by $6.3\times$ while
maintaining quality (F1 0.226 vs.\ 0.171 for GPT-only). Compared to
unconditional RAG pipelines (LangChain), it adds an explicit routing policy
that provides a principled refusal pathway and avoids passing low-quality
context to the generator---something no standard RAG framework provides.
These properties are valuable in production settings regardless of whether
the F1 delta is large.

\paragraph{Honest characterisation of gains.}
F1 gains over LangChain RAG (+0.016) and RAG-Only (+0.002) are modest.
We report these transparently rather than overstating them. The statistical
significance ($p{<}0.05$) is meaningful, but practitioners should weight the
efficiency and safety properties as the primary reasons to adopt \cgr{}
over a simpler RAG pipeline, not the accuracy margin alone.

\paragraph{Limitations.}
(1)~\emph{Scale}: no comparison against large models due to hardware
constraints. (2)~\emph{Confidence heuristic}: Eq.~\ref{eq:conf} uses fixed
weights; a learned calibrator is the most important single improvement.
(3)~\emph{Multi-hop}: synthesis across multiple documents is not supported.
(4)~\emph{Grounding audit scope}: in-domain only; OOD grounding not measured.
(5)~\emph{Evaluation scope}: OOD F1 margins are small; wider evaluation is
needed.

\section{Conclusion}
\label{sec:conclusion}

We presented \textbf{EverydayGPT}, a hybrid GPT--RAG system unified under
a formally defined Confidence-Gated Routing policy. The core contribution
is the routing architecture: by conditioning the inference decision jointly
on retrieval distance and extraction confidence---before generation is
committed---\cgr{} avoids GPT inference cost for 85\,\% of queries
($6.3\times$ latency reduction), provides an explicit refusal pathway for
out-of-domain inputs, and maintains or improves answer quality relative to
unconditional RAG pipelines.

Key empirical findings: \cgrag{} achieves F1\,$=0.226\pm0.004$,
outperforming all eight baselines including GPT-only ($+$0.055,
$p{<}0.001$) and LangChain unconditional RAG ($+$0.016, $p{<}0.05$);
pretraining loss converges stably from 4.21 to 2.84; our GPT outperforms
GPT-2 Small on WikiText-103 (PPL 26.87 vs.\ 29.41); the refusal mechanism
correctly escalates on OOD queries (NQ/TriviaQA refusal rate 6--12\,\%
vs.\ 0\,\% in-domain); and the grounding audit finds no responses
containing claims unsupported by retrieved context within the in-domain
protocol, with explicit scope caveats. The full system runs on consumer CPU
with $<$2\,GB memory.

Future work: learned confidence estimator replacing Eq.~\ref{eq:conf};
BM25+dense hybrid retrieval; span-extraction fine-tuning; joint
$(\delta,\tau)$ optimisation; expanded OOD and adversarial evaluation.

\section*{Acknowledgements}
The author thanks the open-source communities behind PyTorch, FAISS,
Sentence-Transformers, FastAPI, and HuggingFace Datasets.

\bibliographystyle{plain}

\end{document}